\documentclass[twocolumn,amsmath, amssymb, aps, pra,reprint, longbibliography,floatfix,nofootinbib,superscriptaddress,showkeys]{revtex4}

\usepackage[utf8]{inputenc}
\usepackage[urlcolor=blue, colorlinks=true,citecolor=blue]{hyperref}

\usepackage{enumitem}

\usepackage{amsmath}
\usepackage{amsmath,amssymb}
\usepackage{gensymb}
\usepackage{tabularx}

\usepackage{wrapfig}
\usepackage{graphicx}
\usepackage{float}
\usepackage{subfigure}
\usepackage{amssymb}
\usepackage{bbold}
\usepackage{color}
\usepackage{ulem}
\usepackage{comment}

\newcommand{\ket}[1]{\left|#1\right\rangle}

\DeclareMathOperator{\Tr}{Tr} 

\usepackage{tikz}
\usepackage{etoolbox} 
\usepackage{listofitems} 

\tikzset{>=latex} 
\colorlet{myred}{red!80!black}
\colorlet{myblue}{blue!80!black}
\colorlet{mygreen}{green!60!black}
\colorlet{mydarkred}{myred!40!black}
\colorlet{mydarkblue}{myblue!40!black}
\colorlet{mydarkgreen}{mygreen!40!black}
\tikzstyle{node}=[very thick,circle,draw=myblue,minimum size=22,inner sep=0.5,outer sep=0.6]
\tikzstyle{connect}=[->,thick,mydarkblue,shorten >=1]
\tikzset{ 
  node 1/.style={node,mydarkgreen,draw=mygreen,fill=mygreen!25},
  node 2/.style={node,mydarkblue,draw=myblue,fill=myblue!20},
  node 3/.style={node,mydarkred,draw=myred,fill=myred!20},
}

\begin{document}

\preprint{APS/123-QED}

\title{Forecasting steam mass flow in power plants using the parallel hybrid network}

\author{Andrii Kurkin}
\address{Terra Quantum AG, 9000 St.~Gallen, Switzerland}

\author{Jonas Hegemann}
\address{Uniper Technologies GmbH, 45896 Gelsenkirchen, Germany}

\author{Mo Kordzanganeh}
\address{Terra Quantum AG, 9000 St.~Gallen, Switzerland}

\author{Alexey Melnikov}
\thanks{Corresponding author, e-mail: alexey@melnikov.info
\begin{center}
\fbox{
\begin{minipage}{0.45\textwidth}
Please check the published version, which includes all the latest additions and corrections: Eng. Appl. Artif. Intell. 160:111912, 2025, DOI: \href{https://doi.org/10.1016/j.engappai.2025.111912}{10.1016/j.engappai.2025.111912}
\end{minipage}
}
\end{center}
}
\address{Terra Quantum AG, 9000 St.~Gallen, Switzerland}

\begin{abstract}
Efficient and sustainable power generation is a crucial concern in the energy sector. In particular, thermal power plants grapple with accurately predicting steam mass flow, which is crucial for operational efficiency and cost reduction. In this study, we use a parallel hybrid neural network architecture that combines a parametrized quantum circuit and a conventional feed-forward neural network specifically designed for time-series prediction in industrial settings to enhance predictions of steam mass flow 15 minutes into the future. Our results show that the parallel hybrid model outperforms standalone classical and quantum models, achieving more than 5.7 and 4.9 times lower mean squared error loss on the test set after training compared to pure classical and pure quantum networks, respectively. Furthermore, the hybrid model demonstrates smaller relative errors between the ground truth and the model predictions on the test set, up to 2 times better than the pure classical model. These findings contribute to the broader scientific understanding of how integrating quantum and classical machine learning techniques can be applied to real-world challenges faced by the energy sector, ultimately leading to optimized power plant operations. To our knowledge, this study constitutes the first parallel hybrid quantum-classical architecture deployed on a real-world power-plant dataset, illustrating how near-term quantum resources can already augment classical analytics in the energy sector.
\end{abstract}


\maketitle

\section{Introduction}
Accurately forecasting steam‑mass‑flow
15 min ahead, under tight latency and data‑governance constraints, is a
long‑standing bottleneck for fuel‑efficiency optimisation in
waste‑to‑energy power plants. Quantum-enhanced machine learning (or quantum machine learning, QML) has emerged as a rapidly growing field, combining quantum computing with machine learning to develop new models with the potential to revolutionize data analysis \cite{Dunjko2018Machine, Melnikov2023Quantum}. 

Motivated by a industrial use-case, we focus on
short-horizon forecasts of steam mass flow, a key quantity for minimising fuel
consumption, limiting emissions, and preventing
thermal-stress incidents.
The task is technically demanding: the signal combines a smooth
diurnal trend with abrupt perturbations caused by heterogeneous waste
feed; control systems require a new prediction every few seconds; and
strict data-governance rules cap the amount of labeled data that can
be exported for model training.
These characteristics make steam-flow forecasting an ideal
stress-test for near-term quantum machine learning:
\begin{enumerate}[label=(\roman*)]
\item the quasi-periodic component can be captured compactly by the
      truncated Fourier spectra naturally produced by variational quantum
      circuits,
\item the small yet diverse data set aligns with recent theoretical
      bounds showing that parameter-efficient QNNs can generalise
      from fewer examples than comparable classical nets \cite{Caro2022Generalization,Skolik2022Quantum}, and
\item the tight inference-time budget rules out deep autoregressive
      architectures, directing attention toward shallow hybrid
      designs with minimal latency.
\end{enumerate}
Consequently, our central question is whether a noise-robust,
parallel hybrid quantum neural network can already deliver a
measurable accuracy gain on this real-world forecasting problem.
The remainder of the paper answers this question empirically.

A widely adopted approach to QML uses trainable quantum circuits as machine learning models similar to widely known classical neural networks. These circuits consist of encoding and variational unitary gates. The former maps classical data into quantum states, while the latter are parameterized gates trained using classical optimizers to minimize a cost function. The output of the circuit is obtained by measuring an observable on the final state, which produces a vector of real numbers representing the model's prediction. This approach is known in the literature as parametrised quantum circuits (PQCs) \cite{Benedetti2019Parameterized, Jerbi2021Parametrized}, quantum neural networks (QNNs) \cite{Farhi2018Classification, McClean2018Barren, Kordzanganeh2023Exponentially}, variational quantum circuits \cite{Skolik2022Quantum, McClean2016Theory, Romero2021Variational} or quantum circuit learning \cite{Mitarai2018Quantum}. 
The QML toolbox has broadened far beyond variational-ansatz regressors:  
examples include the Hamming-distance quantum $k$-nearest-neighbour algorithm \cite{Li2021Quantum}, the multi-party semi-quantum private-comparison protocol on $d$-dimensional states proposed in \cite{Gong2025Multi}, and the quantum convolutional neural-network architecture \cite{Cong2019Quantum}.  
Acknowledging these developments situates our parallel-hybrid network within the wider effort to translate diverse quantum models to real-world data-science applications such as steam-flow forecasting.

It has been demonstrated on toy datasets that QML models need fewer steps to converge smaller error \cite{Abbas2021Power, Perelshtein2022Practical} and have stronger generalisation ability from fewer data points \cite{Perelshtein2022Practical,Caro2022Generalization} compared to their classical counterparts. However, today, the quantum computing infrastructure cannot yet create competitive quantum models to tackle real-world, ill-structured data science problems \cite{Schuld2022Quantum}. This is only exacerbated by the barren plateau problem, discovered in \cite{McClean2018Barren}, suggesting that large QML models are challenging to train high qubit count. Since NISQ devices limit the freedom in the machine learning model choice, research is instead focused on the hybrid quantum neural networks (HQNN) paradigm -- a combination of classical and quantum models~\cite{Perelshtein2022Practical,Sagingalieva2023HybridCar}. It was shown \cite{Sagingalieva2023Hybrid, Sagingalieva2023HybridCar, Rainjonneau2023Quantum,Senokosov2024Quantum,Sedykh2024Hybrid} that such models can outperform classical counterparts, making this concept attractive for further research.

Classical machine learning techniques have been widely used in the energy sector for various applications, including time series prediction. One common use case is forecasting energy demand or generation to enable effective planning and operation of energy systems \cite{Nassimiha2023Short, Yang2023Generalised, Jiang2023Dayahead, Waterwheel2023Abdelhamid}. However, in this work, we successfully attempted to solve such kind of real-world problem using HQNN. In collaboration with the plant operator we forecast the average
steam-mass-flow 15 min into the future, using exactly the data that are available to the on-site control system. A growing body of work has explored sequential hybrids in which a quantum block feeds into, or receives inputs from, a classical network \cite{Mari2020Transfer,Zhao2019Qdnn,Dou2021Unsupervised,Sebastianelli2021Circuit,Pramanik2021Quantum,Perelshtein2022Practical}. While successful on toy problems, those cascaded designs impose extra latency and create training instabilities when the two branches must be optimised jointly. By contrast, our parallel hybrid network (PHN) processes the same feature vector simultaneously on a shallow variational circuit and a lightweight perceptron, and combines their outputs only at the final layer. This removes the sequential bottleneck, preserves sub‑second inference times that are essential for plant control, and – as recently proven in \cite{Kordzanganeh2023Parallel} – achieves the target accuracy with far fewer parameters than an equivalent sequential hybrid. The present work is therefore not merely a new application, but a first industrial‑scale demonstration of the PHN concept.

Beyond selecting network topologies, practical deployment hinges on tuning hundreds of continuous parameters
(e.g.\ learning-rates\cite{Kingma2017Adam}, weight initialisations\cite{He2015Delving,Glorot2010Understanding})
and discrete hyper-parameters (circuit depth, hidden widths, number of qubits\cite{Bergstra2012Random,Snoek2012Practical}).
Treating those quantities as concurrent objectives leads naturally to multi-objective optimisation (MOO) frameworks that search for
Pareto-optimal trade-offs instead of a single global minimum\cite{Deb2002Fast,Knowles2006ParEGO, Greylag2024Sayed}.
A recent example outside quantum ML is the adaptive NSGA-III with a chaos sequence (NSGA-CS) proposed in \cite{Lei2023Improved},
which jointly sizes power-train components and tunes an on-line energy-management policy while interacting with a high-fidelity vehicle model.
Analogous MOO strategies—e.g.\ evolutionary NSGA variants\cite{Deb2014Evolutionary} or Bayesian Pareto frontiers\cite{Sheikh2022Bayesian}—could be applied to the hyper-parameter space of our
parallel hybrid network to balance forecast error, simulator run-time, and qubit count.
While such an exploration lies beyond the scope of the present study, we agree that it represents a promising avenue for future improvements.

Building on the theoretical proposal of PHN in ~\cite{Kordzanganeh2023Parallel}, this work delivers the first practical instantiation of that architecture on an industrial task. In parallel, we provide a transparent PyTorch + PennyLane reference implementation and accompany the model with Fisher-information, Fourier-accessibility, and ZX-calculus analyses that quantify its trainability and expressivity, offering a reusable diagnostic toolkit for future quantum-classical models. 

The structure of the paper is as follows. Section \ref{sec:problem_arch} provides context for the power plant, specifies the abstractions required to convert this into a data science problem and describes the dataset. Section \ref{sec:quantum_arch} gives the pre-processing steps, baseline classical and hybrid architectures. Then we present the details of our model training and performance results, comparing our hybrid model with the fully quantum and classical counterparts in Section \ref{sec:results}. Our work concludes with discussing our findings in Section \ref{sec:discussion}. Additionally, we offer an analysis of our quantum circuit in the Appendix.

\section{Problem statement and context} \label{sec:problem_arch}

\subsection{Industrial boiler overview} \label{sec:uniper_operaite}
\begin{figure*}[t!]
    \centering
    \includegraphics[width=1\textwidth]{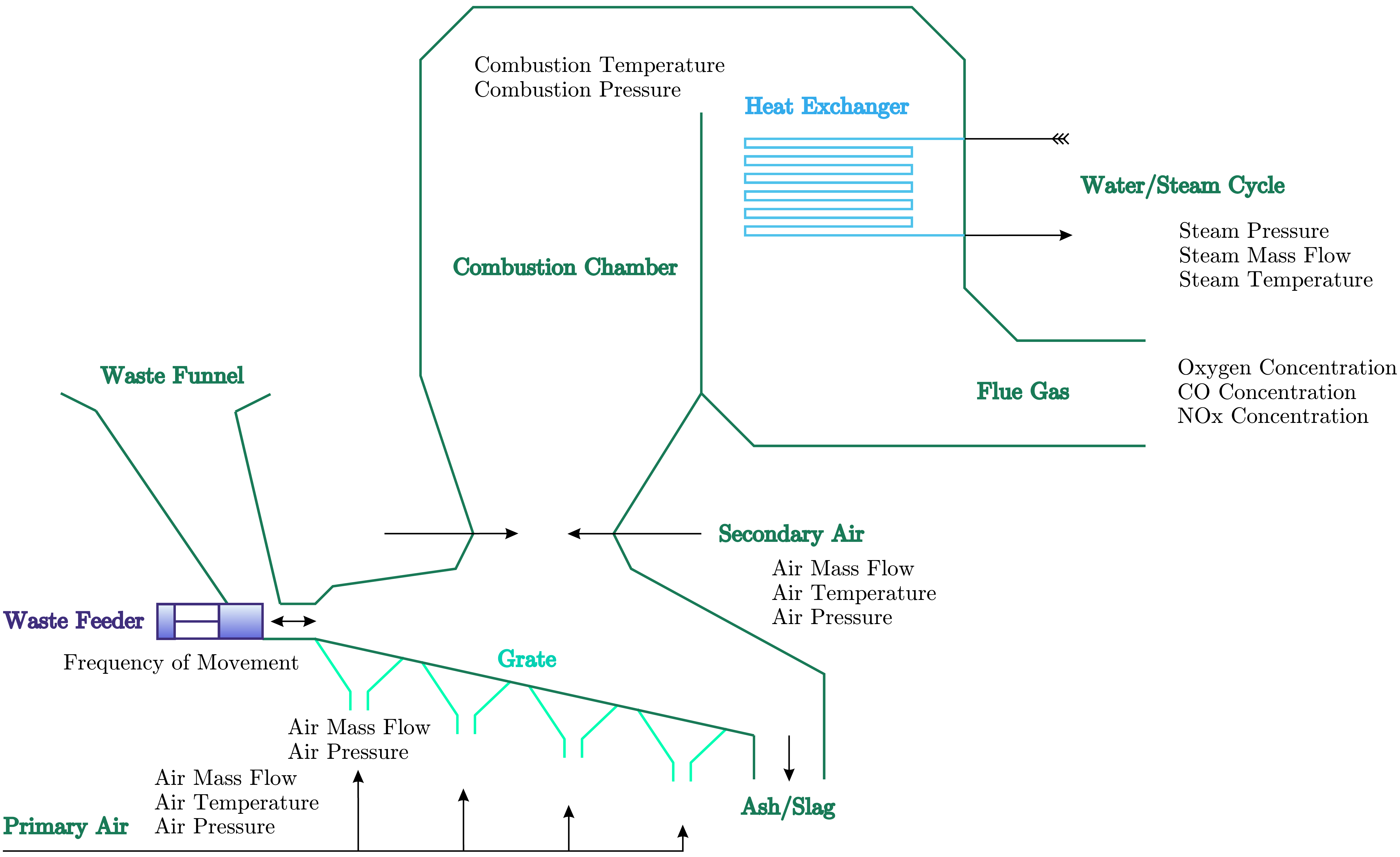}
    \caption{The structure of a typical grate-fired boiler in a waste-to-energy power plant. The waste is dropped into the funnel above the waste feed by a crane, typically a few tons at once. Then, the hydraulic waste feed, which moves back and forth, pushes the waste onto the grate. The grate is divided into a few subsequent zones, which can be moved independently while combustion air enters the chamber from the bottom. The main combustion typically occurs in the second and third zone on the grate, which are in the middle of the grate. The resulting flue gases stream upwards and then along the predefined way until the chimney is reached, where flue gases exit into the atmosphere. On the way, the flue gases pass by the heat exchange mechanisms (pipes), transferring heat energy to the water steam cycle. Eventually, superheated steam is fed into a steam turbine to generate electricity. Measurements are placed all over the boiler and produce time series data around the clock. This data is used to monitor and control the process but can likewise be used to build machine learning models, as we do in this paper.}
    \label{fig:power_plant}
\end{figure*}

A thermal power plant operates by converting heat energy into electricity through a series of processes involving fuel combustion, heat exchange, and steam generation. In waste-to-energy and biomass-to-energy power plants, solid waste or biomass is used as fuel. The combustion process generates hot flue gases, which transfer heat to water, turning it into steam. The superheated steam drives a turbine connected to a generator, which produces electricity. Throughout the power plant, numerous sensors measure parameters such as steam mass flow, air flows, waste feed, temperatures, and chemical concentrations in the flue gas. This data is crucial for monitoring and controlling the plant’s operation and developing machine learning models for forecasting and optimization purposes.

Predicting steam mass flow in a thermal power plant is essential for maintaining operational stability, optimizing energy output, controlling emissions, planning maintenance activities, and reducing costs. Accurate forecasts of steam mass flow enable operators to make informed decisions about adjusting various parameters, such as fuel input and air supply, to ensure smooth and efficient operation. Furthermore, predicting steam mass flow can help in controlling emissions generated by the power plant, minimizing the production of pollutants that negatively impact the environment and public health.

A fully connected feed-forward network is currently deployed on the
plant’s distributed-control system for this purpose. In waste-to-energy and biomass-to-energy power plants, combustion processes are highly dynamic due to volatile fuel quality resulting, e.g. from varying geometric or chemical properties of the typically solid fuel components. This corresponds to continuous stimulation of the combustion process and constantly changes the equilibrium of the combustion process, which is given by a certain ratio between fuel infeed and airflow. Thus, continuous control (of, e.g., the airflow) is needed to maintain a steady and stable operation, which is usually achieved by using CCS (combustion control systems). Various CCSs exist, ranging from simple conventional systems built from combinations of PID (proportional-integral-derivative) controllers over fuzzy or rule-based systems to model predictive control. Most power plants rely on PID controllers, and only a few employ more advanced process control techniques, mainly because implementing them takes a lot of time and is related to high costs. Typical optimisation goals are to raise the average or reduce the variance of the energy output in terms of steam mass flow or electricity generation or to reduce emissions in terms of CO or NOx concentrations in the flue gas. There are two ways to achieve these operational goals, which are (i) using a controlling neural network that directly interacts with control levers and (ii) using a neural network that provides forecasts and serves as an assistant system to help humans manipulate the process proactively. This paper concentrates on improving (ii) by employing hybrid quantum neural network forecasts.

\subsection{Forecasting problem} \label{sec:data_science}

\begin{figure*}
    \centering
    \includegraphics[width=\textwidth]{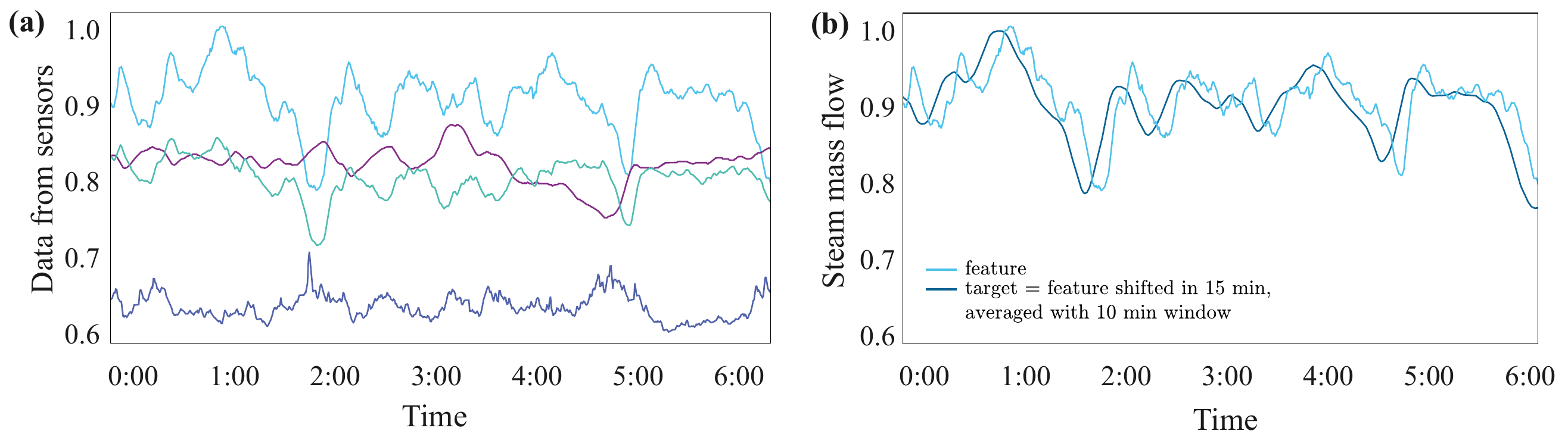}
    \caption{Non-dimensionalized feature and target data used for time series prediction. (a) Features consist of $192$ time series, including steam mass flow, with the first four-time series displayed for clarity over $6$ hours. (b) Non-dimensionalized target data shows steam mass flow values shifted $15$ minutes into the future and averaged over a $10$-minute interval, displayed only for the first sensor for simplicity. Note that the shifted and averaged signal will be used by human operators as an early indicator for problematic process situations. This indication will help them to take proactive actions.}
    \label{fig:train_data}
\end{figure*}

In this section, we present the problem statement and discuss the dataset from the power plant for forecasting problems. The objective is to develop a model that can predict the steam mass flow for two sensors $15$ minutes into the future based on the current values of all the power plant parameters, including air flows, waste feed, temperatures, chemical concentrations in the flue gas, and steam mass flow, among others. Thus, we solve a multivariate regression problem in continuous space.

The prediction timeframe is determined by the characteristic timescales of the combustion process, which are depicted in Fig.~\ref{fig:power_plant}. The combustion process entails several stages, including waste feeding, transport across the grate, flue gas stream, and heat exchange. The airflow needs to be changed continuously in response to fuel quality changes. Assuming a constant fuel quality, the system requires $5$ to $15$ minutes to return to equilibrium \cite{Magnanelli2020Dynamic,Hu2024Efficient,Astroem2000Drum}. Consequently, we deemed it appropriate to consider a prediction horizon of $15$ minutes for this task. Thus, the forecasted timeframe should be sufficient to take appropriate action early enough.

Feature and target data used in our study are represented in Fig.~\ref{fig:train_data}. The dataset consists of $192$ time series, which are our features, each representing measurements of power plant parameters captured by various sensors throughout the facility, also including numerical derivatives and moving averages for each sensor. Concretely, the 192 raw channels span five broad categories: (i) steam-cycle quantities such as pressure and mass flow, (ii) primary- and secondary-air flows plus associated damper positions, (iii) flue-gas composition and temperature measurements, (iv) waste-feed actuation and bunker-level signals, and (v) internal set-points and helper variables generated by the combustion-control system. The targets we aim to predict are represented by two time series. Specifically, each target variable corresponds to the steam mass flow value for a given sensor that has been shifted $15$ minutes into the future and averaged over a $10$-minute window. This averaging process is intended to create a more uniform steam mass flow signal and reduce fluctuations. Our training dataset consists of $6,500$ timestamps, which only represent $1\%$ of the original dataset used to train the classical model currently in production. We chose to reduce the dataset due to the time-consuming nature of training a quantum neural network using existing simulators\cite{Kordzanganeh2023Benchmarking}. Recent theoretical analyses show that suitably parameter-efficient quantum neural networks can enjoy a polynomial reduction in sample complexity relative to comparable classical models \cite{Caro2022Generalization}, indicating that strong generalisation from a small yet diverse subset is plausible. Because our ansatz satisfies this parameter-efficiency condition, the bound implies that a few thousand well-chosen examples already push the statistical error below the shot-noise floor, so further enlarging the dataset would offer only marginal gains while incurring prohibitive simulation time. Therefore, such a simplification can be considered justified. Future work can revisit this limitation by training the
PHN on progressively larger fractions of the 650 k-timestamp archive and
by evaluating cross-plant generalisation on historian data from additional
facilities, potentially leveraging transfer-learning pre-training stages
on classical surrogates. The 192 input channels arise from 24 physical sensor
streams—steam-flow, primary-air actuators, flue-gas composition, waste-feed
logistics and internal control helpers—each augmented with eight derived
quantities (first/second numerical derivatives plus six windowed moving
averages).  All raw values are z-scored; in the training split this
produces a mean of $0.00\!\pm\!0.01$ and variance of $1.00\!\pm\!0.03$,
and the held-out test split matches these first-order statistics within the
stated uncertainties, confirming consistent scaling.

\section{Machine learning models} \label{sec:quantum_arch}

\begin{figure*}
    \centering
    \includegraphics[width=\textwidth]{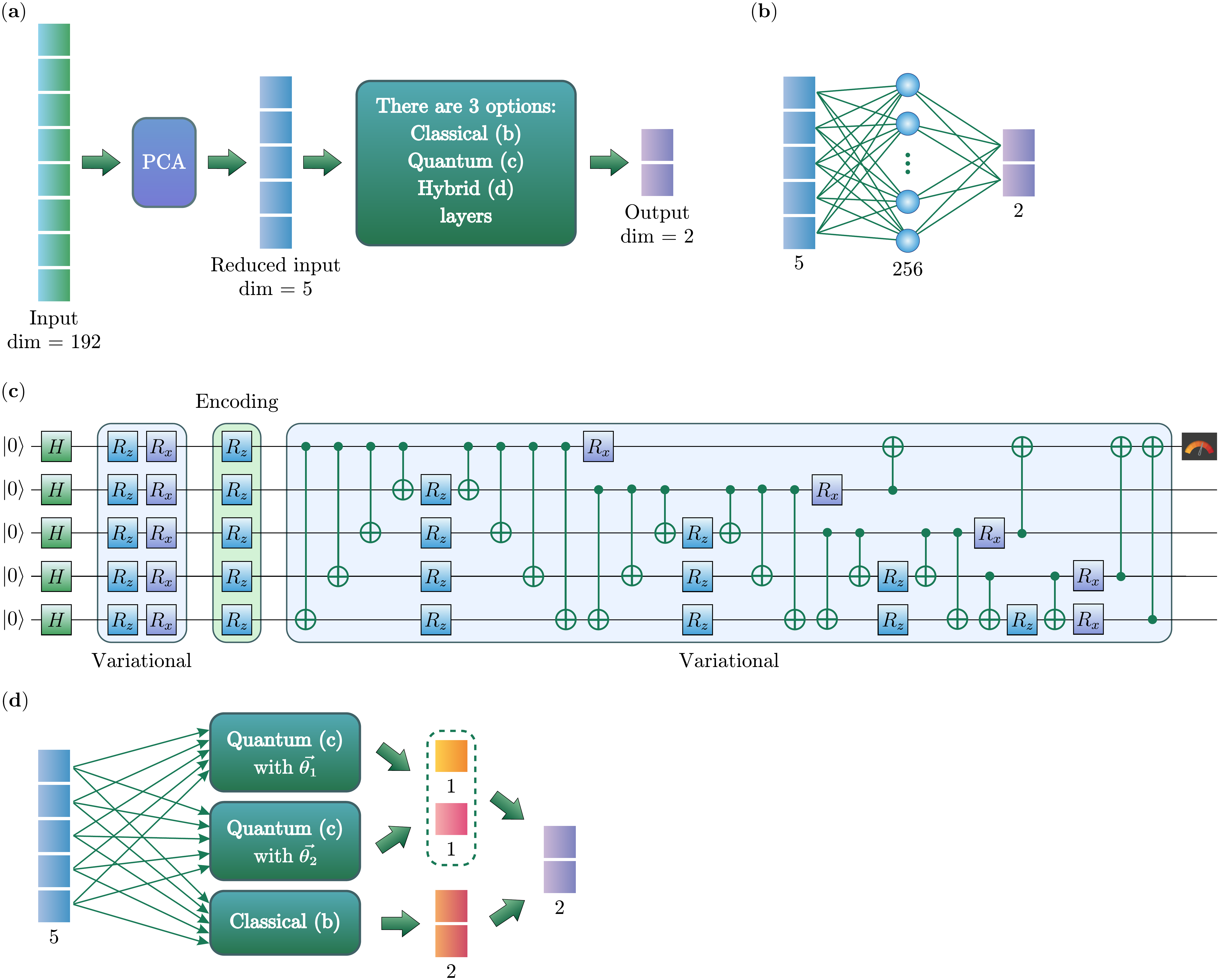}
    \caption{(a) Problem definition and general pipeline of solution. The task is to predict steam mass flow for two sensors at time $t + 15$ min, given values of multiple time series (192) from different sensors at time $t$. The input vector $\mathbf{x}$ is pre-processed with PCA to reduce the feature space from $192$ to $5$. Then, an appropriate machine learning algorithm (classical, quantum, or hybrid) is applied to reduced input to predict the output, which is a $2$-dimensional vector corresponding to the target value for each of the two sensors. (b) Classical neural network architecture. The architecture consists of $5$ input neurons, $256$ hidden neurons, and $2$ output neurons. (c) Parameterized quantum circuit (PQC). The quantum circuit includes $5$ qubits and a set of encoding and variational gates. The first qubit is measured in the $Z$-basis to yield a real-valued output. (d) shows the parallel hybrid network architecture. The architecture consists of a classical neural network and two identical PQCs that operate independently. Each of them takes a $5$-dimensional reduced input. The classical and quantum predictions are added to the corresponding components, resulting in the final output.}
    \label{fig: Problem solution & architectures}
\end{figure*}

\subsection{Pre-processing} 
An effective pre-processing of data is crucial for any machine learning pipeline, especially when dealing with quantum neural networks. The scarcity of qubits on physical quantum computers and the high computational cost of PQCs simulations are significant challenges in this field \cite{Kordzanganeh2023Benchmarking, Tang2022Cutting}. However, even small quantum networks combined with classical ones can still achieve supremacy in some cases. It is important to transform the data into a space with dimensions suitable for quantum processing to overcome this challenge. In our approach, we use principal component analysis (PCA), a powerful data analysis technique that reduces the dimensions of the original dataset \cite{Hill2016PythonML}. By projecting the data onto the five main components, we reduce the input vector dimension from $192$ to $5$ while retaining more than $60\%$ of the original data's contribution share as illustrated in Fig.~\ref{fig: Results}(a).  It’s important to note that, in some domains, authors retain enough principal components to explain $80\%$ or more of the variance \cite{Lei2025Theory}. In the present study we deliberately stop at five components (approximately $60\%$) for two reasons: (i) the quantum sub-network provides exactly five data qubits, so additional components would violate the qubit budget or require sequential batching that conflicts with the fully parallel design, and (ii) beyond the fifth component the eigen-spectrum of the covariance matrix flattens and mostly captures sensor noise rather than meaningful dynamics, a practice also advocated in industrial PCA literature \cite{Kourti1995Process}. While applications with richer, less correlated signals might indeed benefit from the $80\%$ rule, we found $60\%$ to be an acceptable trade-off between information content and model complexity for steam-flow forecasting.

After applying PCA, we standardize and normalize the data, which typically has varying scales. These pre-processing steps prepare the data for processing with PQC, for which appropriate architectures will be introduced below.

\subsection{Classical baseline architecture} \label{sec:existing_model}
A classical baseline architecture was utilised to enable future comparisons with hybrid solutions. This architecture is a fully connected neural network with one hidden layer, consisting of $5$ input neurons, $256$ hidden neurons, and $2$ output neurons, as illustrated in Fig.~\ref{fig: Problem solution & architectures}(b). It is important to note that the design of the baseline model was inspired by an existing model currently in production, which addresses a similar problem of multivariate regression. However, the difference lies in the number of targets, as the production model predicts more parameters for various sensors, while our model only predicts steam mass flow for two sensors. The architecture of the production model is also a conventional feed-forward neural network with one hidden layer.

Although the forecasting problem can be effectively solved using recurrent neural networks, such as LSTM models \cite{Schmidt2019Recurrent,Sherstinsky2020Fundamentals} or more advanced Transformer-based architectures \cite{Vaswani2017Attention}, conventional feed-forward neural networks with a more straightforward structure are generally easier to train and faster to process. Conventional neural networks have the advantage that the forward pass, which is necessary for the production environment, can be efficiently implemented on a controller compatible with the distributed control system in a power plant. In addition, conventional feed-forward neural networks provide better explainability compared to more advanced network topologies. For this reason, we used a conventional feed-forward neural network for our baseline architecture in this research. Recent work has shown that recurrent Long Short-Term Memory (LSTM) networks and, more recently, Transformer encoders achieve state-of-the-art accuracy on many industrial forecasting tasks, including energy-system optimisation \cite{Lei2024Physics}.  
Nevertheless, these alternatives come with different computational trade-offs in an online setting.  
Transformers require pair-wise self-attention over the input sequence, giving an $\mathcal{O}(T^{2})$ complexity in the sequence length $T$, whereas both MLPs and LSTMs scale linearly.  
For a power-plant controller that must issue a prediction every few seconds, the quadratic cost is prohibitive.  
LSTMs are more attractive, and we fully agree with the reviewer that an LSTM baseline would be informative.  
Importantly, the choice “LSTM vs.\ MLP’’ is orthogonal to “hybrid vs.\ classical’’: an LSTM branch could be embedded in our parallel hybrid network in exactly the same way as the current MLP, and a quantum-LSTM hybrid is an interesting avenue for follow-up work.  
We leave that exploration to future work and focus here on isolating the quantum–classical split, which is the main contribution of this paper.

\subsection{Hybrid quantum-classical architecture}\label{sec:HQNN} 
A good understanding of the data structure is crucial in building an effective architecture for problem-solving. In our case, our goal is to predict the values of a time series at a given time. By analyzing Fig.~\ref{fig:train_data}, we can observe that the prediction follows a sinusoidal pattern with some irregularities. It is known that a classical neural network with one hidden layer is an asymptotically universal approximator \cite{Schuld2021Effect,Skolik2021Layerwise}
. This was also shown for parameterised quantum circuits (PQC) - quantum neural networks capable of universal approximation \cite{Schuld2021Effect}. However, PQCs achieve this by fitting a truncated Fourier series over the samples. With this in mind, we use the parallel hybrid network (PHN) configuration introduced in \cite{Kordzanganeh2023Parallel}, which differs from the previous sequential quantum-classical hybrid models \cite{Mari2020Transfer,Zhao2019Qdnn,Dou2021Unsupervised,Sebastianelli2021Circuit, Pramanik2021Quantum, Perelshtein2022Practical, Sagingalieva2023Hybrid, Sagingalieva2023HybridCar,Rainjonneau2023Quantum}. Here, the quantum and classical parts process the data independently and simultaneously without interfering with each other. The quantum circuit approximates the sinusoidal part, while the classical network fits the protruding sections. Finally, the predictions from both parts are combined to obtain the final prediction. This approach lets the classical model only adjust its weights and does not interfere with the quantum circuit during the training procedure.

A PHN architecture allows the truncated Fourier series produced by the VQC and the non-harmonic residual produced by the perceptron to coexist and be combined linearly. Reference \cite{Kordzanganeh2023Parallel} formally proves this property and shows that, on periodic-plus-spikes data, a PHN attains the target accuracy with far fewer parameters than an otherwise identical sequential hybrid. Because the steam-flow signal likewise contains a smooth periodic backbone with local irregularities, we adopt the parallel topology for the present study. Readers interested in the full derivation, including the
Fourier–coeff\-icient bounds and the associated sample-complexity result, are referred to Section II of ~\cite{Kordzanganeh2023Parallel}.

The proposed architecture comprises two identical parameterized quantum ansatz circuits, which will be introduced below, and a classical fully-connected neural network, discussed in Section \ref{sec:existing_model} and illustrated in Figure \ref{fig: Problem solution & architectures}(b). The complete PHN architecture is depicted in Figure \ref{fig: Problem solution & architectures}(d), with the same feature vector as input to both the classical and quantum circuits. Before this, the input feature vector undergoes a PCA procedure to reduce its dimensions from 192 to 5.

The quantum layer (PQC) depicted in Figure \ref{fig: Problem solution & architectures}(c) consists of five qubits. The layer begins with applying a Hadamard transform to each qubit, followed by a sequence of variational gates consisting of rotations along the $z$ and $x$ axes for each qubit. The reduced input data (with a dimensionality of 5) is then embedded into the rotation angles along the $z$-axis. Subsequently, another variational block is applied, consisting of a sequence of $R_{ZZ}$ gates that alternate periodically with rotations along the $x$-axis and conventional CNOT gates. The complete gate sequence can be found in Figure \ref{fig: Problem solution & architectures}(c). Finally, the local expectation value of the Z operator is measured for the first qubit, producing a classical output suitable for additional post-processing. In the Appendix, we analyze a simplified version of this quantum circuit using three approaches to assess its efficiency, trainability, and expressivity.

All quantum layers are programmed in PennyLane (v0.32) and executed on
the \texttt{lightning.qubit} back-end in analytic mode
(\texttt{shots}=None), so every forward pass returns an exact
expectation value.  Gradients are obtained with the adjoint
differentiation method and passed to PyTorch through
\texttt{qml.qnn.TorchLayer}.  Each variational circuit (Fig.~\ref{fig: Problem solution & architectures}c) acts on five data qubits. It begins with a layer of Hadamard gates, followed by a shallow variational layer composed of $R_Z$ and $R_X$ rotations. The data are then encoded via an $R_Z$ rotation, after which the circuit applies repeated blocks of parameterised single-qubit $R_Z$ gates interleaved with entangling CNOT gates.
Optimisation uses Adam with learning rates of $10^{-3}$ for quantum
parameters and $10^{-4}$ for classical parameters. 

The predictions of the classical network and the quantum circuits are combined to generate the final prediction. Specifically, the output of the first quantum circuit is added to the first component of the classical network's prediction vector. In contrast, the output of the second quantum circuit is added to the second component of the vector. This results in the final classical-quantum output, which has the potential to enhance accuracy and efficiency for time series prediction.

\section{Training and results} 
\label{sec:results}
\begin{figure*}
    \centering
    \includegraphics[width=\textwidth]{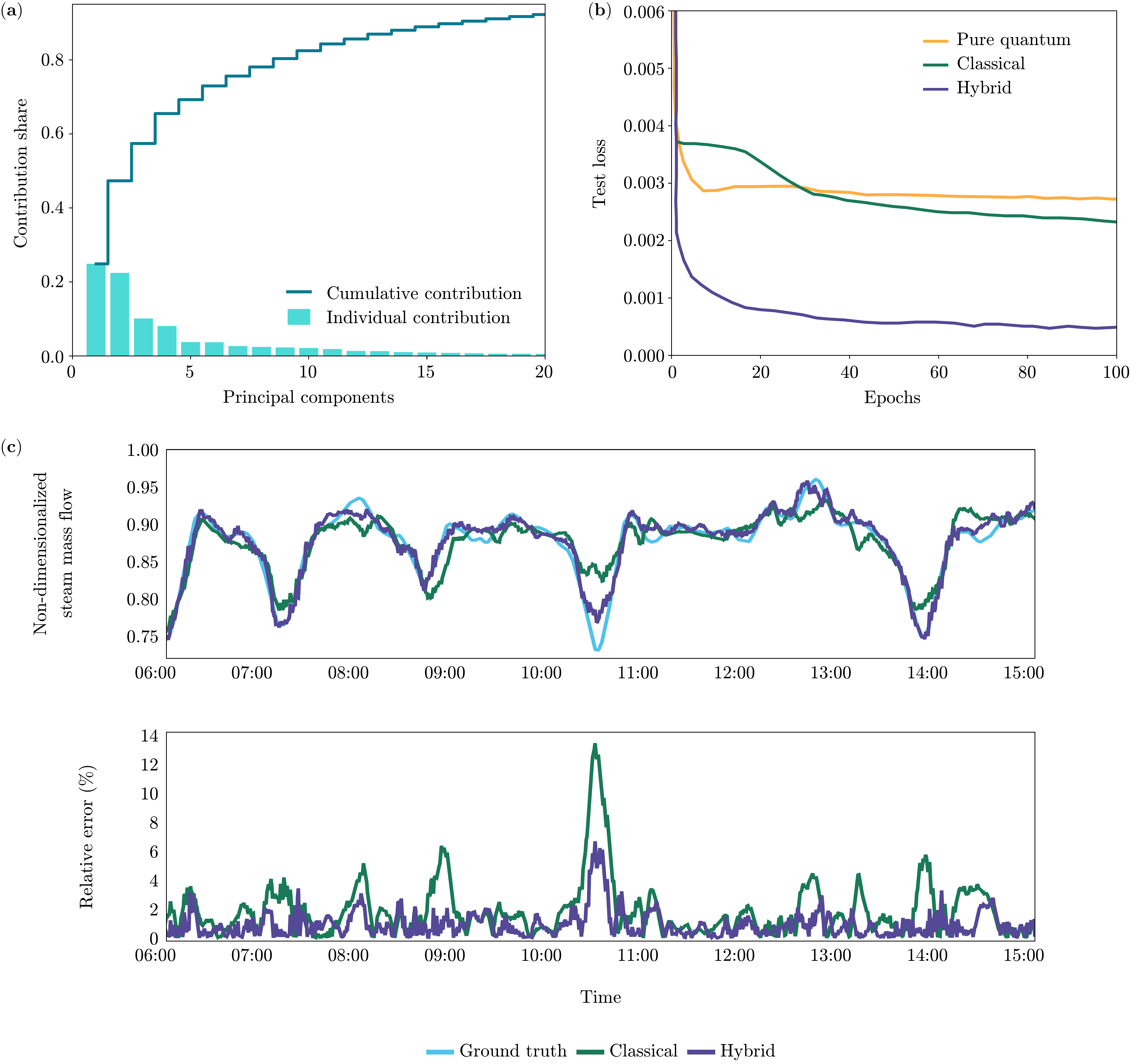}
    \caption{(a) The results of the principal component analysis (PCA). We can see that the first five principal components cover a significant majority of the contribution share. (b) Loss performance comparison between the hybrid, pure classical, and pure quantum models.  We see that while they each perform sub-optimally separately, classical and quantum networks can form a hybrid network that outperforms both by a significant margin. (c) The fit of the classical and hybrid circuit to the time series in the test region.  The top graph shows the classical and hybrid predictions on unseen data. In contrast, the bottom graph shows its residuals -- relative errors between ground truth and the model prediction at each point.}
    \label{fig: Results}
\end{figure*}

When using this architecture, one must take extra caution when tuning the hyper-parameters. This arises due to the separability of the three parallelized architectures. It is important to make sure none of the 3 fully dominates the training, resulting in a model stuck in a local minimum\footnote{In some cases, it could be plausible that the global minimum is where only one network dominates and that no contribution is made by the other two, but this is unlikely as for the most part the quantum and classical networks can produce values independently from each others' parameters. This is especially unlikely in the case of the dataset in Fig.~\ref{fig:train_data} due to its periodicity.}.  For this reason, we make sure that the quantum network has the chance to train first to create a sinusoidal landscape, and then the classical network begins to contribute.  We do this by reducing the learning rate of the classical parameters compared with the quantum ones.  This ensures that the quantum networks can train and fit a sinusoidal function before the classical make any real contribution.  At some point in the training, the quantum network has achieved a minimum, and its gradient values are so small that the classical network begins its meaningful training stage. 

All machine learning experiments were conducted on the QMware cloud platform \cite{Qmw2022First}. PyTorch library \cite{Paszke2019Advances} was used to implement the classical part, while the quantum one was implemented using the PennyLane framework \cite{Bergholm2022Automatic}. We used the \texttt{lightning.qubit} device, which implements a high-performance C++ backend. The standard backpropagation algorithm was applied to the classical part of our hybrid quantum neural network to calculate the loss function's gradients for each parameter. In contrast, the adjoint method was employed for the quantum part.

In Fig.~\ref{fig: Results}(b), the loss performance is presented for the hybrid architecture compared to the purely classical and purely quantum networks. The pure classical network exclusively employs the classical neural network, whereas the pure quantum network utilizes only the two quantum models in the absence of classical components. It is evident from the results that the hybrid architecture outperforms the pure classical and pure quantum counterparts by a significant margin. Specifically, after 100 epochs of training, the mean squared error (MSE) loss for the hybrid architecture is more than 5.7 and 4.9 times lower than that of the purely classical and purely quantum networks. This confirms that the gain arises from the combined action of the two branches rather than from the quantum circuit alone. At epoch 100 the PHN reaches an MSE of 0.018 on the
training set and 0.019 on the validation set, a ratio of 1.06, which
shows no evidence of over-fitting.  Hyper-parameters were selected by a
coarse grid search against this same validation split. Table~\ref{tab:ablation} reports the resulting test-set
MSE relative to the PHN.

\begin{table}[ht]
    \caption{Test-set MSE for the ablation baselines.
             Values are shown relative to the PHN (\(\text{MSE}=1.0\)).}
    \label{tab:ablation}
    \centering
    \begin{ruledtabular}
    \begin{tabular}{lcc}
        \textbf{Model} & \textbf{Constituents} & \textbf{MSE (×\,PHN)} \\ \hline
        Classical only & MLP (256\,$\rightarrow$\,2)          & 5.7 \\ 
        Quantum only   & Two identical VQCs                  & 4.9 \\ 
        \textbf{PHN (ours)}     & MLP + VQC in parallel               & 1.0 \\ 
    \end{tabular}
    \end{ruledtabular}
\end{table}

Fig.~\ref{fig: Results}(c) shows the fit of the classical and hybrid circuits to the time-series data in the test region. The top graph displays the classical and hybrid predictions on unseen data. In contrast, the bottom graph depicts their residuals, representing the relative errors between the ground truth and the model predictions at each point. Overall, the hybrid model predictions are much closer to the ground truth with a smaller relative error, up to 2 times lower than the classical approach. Therefore, a PHN-based approach is the most efficient strategy in this case. 

\section{Discussion}
\label{sec:discussion}
This study has highlighted the potential of quantum machine learning for time series prediction in the energy sector through a novel parallel hybrid architecture. We have demonstrated lower test-set error than traditional classical and quantum architectures by combining independent classical and quantum neural networks. Our approach enables the classical and quantum networks to operate independently during the training, preventing interference between the two.

All training and evaluation in this work were performed with the noise–free \texttt{lightning.qubit} backend; the purpose was to quantify the benefit that quantum resources can bring to steam-flow forecasting under ideal conditions. The variational circuits employed are shallow, a depth for which error–mitigation techniques such as zero-noise extrapolation and probabilistic error cancellation have been shown to restore near-logical accuracy on current NISQ processors \cite{Temme2017Error,Kandala2019Error}. Beyond mitigation, several studies have shown that shallow, over-parameterised variational circuits can adapt their parameters to compensate for moderate hardware noise, thereby preserving accuracy well into the NISQ regime \cite{Rabinovich2024Robustness,Martin2024Effects}. Large-scale simulations of hybrid QCNN-style ansätze under realistic bit-flip and phase-flip channels further corroborate this effect, reporting virtually unchanged test accuracy up to noise probabilities of a few percent \cite{Ahmed2025Noisy}.
A systematic noise sweep is therefore an important, but orthogonal, follow-up study.

Our results reveal that the parallel hybrid model outperforms pure classical and pure quantum networks, exhibiting more than 5.7 and 4.9 times lower MSE loss on the test set after training. Furthermore, the hybrid model demonstrates more minor relative errors between the ground truth and the model predictions on the test set, up to 2 times better than the pure classical model.

These findings suggest that quantum machine learning can be valuable for solving real-world problems in the energy sector and beyond. Future research could explore applying the parallel hybrid quantum neural network approach to other machine learning problems while also increasing the complexity and performance of the model.

\section{Conclusion}\label{sec:conclusion}

This work delivers the first {\it in-situ} demonstration that a
PHN can add measurable value to an
industrial forecasting task under the tight data-governance and
latency constraints typical of power-plant control.
Using only five data qubits and a single hidden layer, the PHN
reduced the test-set mean-squared error by more than a factor of~5.7
relative to a tuned classical baseline, and by a factor of~4.9
relative to a stand-alone variational circuit.  
These empirical gains are consistent with the recently proven
theory \cite{Kordzanganeh2023Parallel}: the quantum branch supplies a
low-frequency Fourier scaffold, while the perceptron branch patches
non-harmonic residuals, yielding a parameter-efficient universal
approximator for “periodic-plus-spikes’’ signals such as steam-flow
time series.

For operators this translates into smoother boiler
operation, tangible fuel savings and reduced emissions, directly
benefiting the waste‑to‑energy sector.  
Because the PHN is data‑agnostic, any industrial process with
quasi‑periodic sensor traces—such as steelmaking furnaces, chemical
reactors or pulp‑and‑paper digesters—can adopt the same template by
re‑training on local historian data.

From an application perspective, the PHN’s shallow depth and modest
qubit count make a near-term hardware deployment plausible, provided
standard error-mitigation techniques (e.g.\ zero-noise extrapolation)
are applied.  The open-source PyTorch + PennyLane implementation
supplied with this paper is therefore meant to serve as a ready
blue-print for follow-up field trials.

Several limitations remain.  First, all training and inference were
performed with an ideal (noise-free) statevector simulator; a
systematic hardware noise study is under way.  Second, the classical
branch used a feed-forward network; replacing it with a
latency-optimised LSTM or transformer may may reduce the remaining error further accuracy.
Third, multi-objective hyper-parameter tuning—balancing error,
run-time and qubit budget—has not yet been explored.

In a prospective roll-out the PHN would operate as an
advisory layer, with confidence thresholds and automatic fall-back to
the incumbent PID logic ensuring safety, transparency and operator
accountability. Because a forward pass (\,\(\approx 40\,\mu\text{s}\) classical \(+\)
\(< 1\,\text{ms}\) remote quantum call) fits comfortably within the
sub-second cycle of modern combustion-control systems, integration is
technically feasible via a gRPC wrapper around the existing prediction
block.

In summary, the present study provides both a quantitative benchmark
and a transparent methodological template for applying hybrid quantum
machine learning to real-world energy-system data.  We hope it will
encourage wider testing of PHNs on other industrial time-series
problems and accelerate the path towards quantum-enhanced, sustainable
power-plant operations.

\section{Acknowledgements}

The authors used OpenAI GPT  exclusively for language polishing after all technical content had been produced; no part of the scientific substance was
generated by AI.

\bibliography{lib}
\bibliographystyle{unsrt}

\onecolumngrid
\appendix

\section*{APPENDIX}
\section*{Quantum Circuit Analysis}\label{sec:appendix}
In this section, we thoroughly examine a parameterized quantum circuit (PQC) that is represented in Fig.~\ref{fig: analysis}(a). This PQC is a 2-qubit toy version\footnote{Carrying out such an analysis for the original architecture can be highly computationally expensive in terms of the Fisher information matrix calculation. It is also not as visual as the Fourier accessibility demonstration. Hence, we decided to analyze a simplified version.} introduced in Section \ref{sec:HQNN}, inheriting its fundamental properties and concepts. Our analysis focuses on three different approaches: the ZX-calculus \cite{Coecke2011Interacting} to examine circuit-reducibility, the Fisher information \cite{Abbas2021Power} to evaluate the trainable parameters and the circuit expressiveness, and the Fourier accessibility \cite{Schuld2021Effect} to investigate the encoding.

\subsection{ZX-calculus}\label{sec:appendix_ZX}
The ZX-calculus is a graphical language initially based on Category Theory that can simplify a quantum circuit to a simpler, equivalent one \cite{Coecke2011Interacting}. It involves transforming the circuit into a ZX graph and applying the ZX-calculus rules introduced in Ref.~\cite{Wetering2020Zx} to reduce the graph to a more fundamental version. After that, the obtained version is mapped again to the quantum circuit. This process results in a new, streamlined circuit that achieves the maximum potential of trainable layers while avoiding fully redundant parameters. If a circuit cannot be further reduced, it is called ZX-irreducible.

In this study, we present a novel quantum circuit that generates a non-commuting graph which cannot be simplified using ZX-calculus rules. This is illustrated in Fig.~\ref{fig: analysis}(b), where adjacent dots are colored differently. Notably, the X-spiders (red dots) appear only at the first wire's end, and the measurement is performed in the Z-basis (green family). Consequently, no pairs of dots can commute with each other, and none of them can fuse. This implies that our circuit has no redundant parameters.

Remarkably, our circuit is designed only to measure the first qubit. At the same time, the encoding and variational parameters of the second qubit (in the original circuit with the other four qubits) heavily influence the measurement outcome of the first qubit. This is achieved using the $R_{ZZ}$ gate, namely parameter $\alpha_5$ introduces pairwise correlations between the $x_1$ and $x_2$ features (in the original circuit, such correlations create between any pair of qubits). As a result, we obtain a highly complex approximator that depends on the features in a non-trivial manner. The subsequent sections present a detailed analysis of such a quantum approximator's qualities.

\subsection{Fisher information}\label{sec:appendix_fim}
Any neural network, classical or quantum, can be considered a statistical model. The Fisher information estimates the knowledge gained by a particular parameterization of such a statistical model. In supervised machine learning, we are given a set of data pairs $(\mathbf{x},y)$ from a training subset and a parameterized model $h_{\boldsymbol{\theta}}(\mathbf{x})$ that maps input data $\mathbf{x}$ to output $y$. The parameterized models family can be fully described by the joint probability of features and targets: $\mathcal{F} :=\{P(\mathbf{x}, \mathbf{y}|\boldsymbol{\theta}): \boldsymbol{\theta} \in  \boldsymbol{\Theta}\}$ and during the training procedure, we want to maximize likelihood to determine the parameters $\boldsymbol{\hat{\theta}} \in \boldsymbol{\Theta}$ for which the observed data have the highest joint probability. We can think of $\mathcal{F}$ as some Riemannian manifold, and the Fisher information matrix can be naturally defined as a metric over this manifold \cite{Amari1998Natural, Abbas2021Power}:

\begin{equation}\label{fishereqn}
F(\boldsymbol{\theta}) = \mathbb{E}_{\{(\mathbf{x}, \mathbf{y}) \sim p\}}[\boldsymbol{\nabla}_{\boldsymbol{\theta}} \log p(\mathbf{x}, \mathbf{y}| \boldsymbol{\theta}) \cdot \boldsymbol{\nabla}_{\boldsymbol{\theta}} \log p(\mathbf{x}, \mathbf{y}| \boldsymbol{\theta})^T]
\end{equation}

According to the findings presented in Ref.~\cite{Abbas2021Power}, when the number of qubits in a model increases, a Fisher information spectrum with a higher concentration of eigenvalues approaching zero indicates that the model potentially suffers from a barren plateau. On the other hand, if the Fisher information spectrum is not concentrated around zero, it is less likely for the model to experience a barren plateau.

Using the Fisher information matrix, we can also describe model capacity: quantifying the class of functions, a model can fit, in other words, a measure of the model's complexity. For this purpose, the notion of effective dimension, firstly introduced in Ref.~\cite{Berezniuk2020Scaledependent} and modified in Ref.~\cite{Abbas2021Power}, can be used:

\begin{equation}
    d_{\gamma,n}(\mathcal{M}_\Theta):=2\frac{\text{log}\left(\frac{1}{V_\Theta}\int_\Theta\sqrt{\text{det}\left(id_d+\frac{\gamma n}{2\pi \text{log} n}\hat{F}(\boldsymbol{\theta})\right)}d\boldsymbol{\theta}\right)}{\text{log}\left(\frac{\gamma n}{2\pi \text{log} n}\right)},
\end{equation}
where $V_\Theta := \int_\Theta d\theta $ is the volume of the parameter space, $\gamma$ is some constant factor \cite{Abbas2021Power}, and $\hat{F}(\boldsymbol{\theta})$ is the normalised Fisher matrix defined as
\begin{equation}
\label{normalisedfisher}
    \hat{F}_{ij}(\boldsymbol{\theta}):=d\frac{V_\Theta}{\int_\Theta \Tr(F(\boldsymbol{\theta}))d\boldsymbol{\theta} } F_{ij}(\boldsymbol{\theta}).
\end{equation}

We calculate the Fisher information for three specific toy circuit configurations: with $N=1$  last trainable layer repetition - contains $7$ trainable parameters, when $N=2$ and $N=3$ repetitions that consist of $10$ and $13$ trainable parameters accordingly. For a finite number of data, taking into account definition (\ref{fishereqn}), the Fisher information estimate with some simplifications can be rewritten as follows:

\begin{equation}
    F(\boldsymbol{\theta}) = \sum_{(\mathbf{x}, \mathbf{y}) \in X \times Y} 
    \frac{\boldsymbol{\nabla}_{\boldsymbol{\theta}} P(\mathbf{x}, \mathbf{y}| \boldsymbol{\theta}) \cdot \boldsymbol{\nabla}_{\boldsymbol{\theta}} P(\mathbf{x}, \mathbf{y}| \boldsymbol{\theta})^T}{P(\mathbf{x}, \mathbf{y}| \boldsymbol{\theta})},
\end{equation}
where the joint probability for QNN can be defined as the overlap between model output and these states:
\begin{equation}
    P(\mathbf{x}, \mathbf{y}|\boldsymbol{\theta}) = 
    \Tr (\rho(\boldsymbol{\theta},\mathbf{x}) \cdot \mathbf{y} \mathbf{y}^{\dagger}).
\end{equation}
 Following Ref.~\cite{Abbas2021Power}, we used $1000$ features samples, each of them comes from Gaussian distribution $\mathbf{x_i} \sim \mathcal{N} (\mu = 0, \sigma^2 = 1)$, and target as specific resultant state $\mathbf{y} \in Y =\{\ket{00}, \ket{01}, \ket{10}, \ket{11}\}$ which are all possible basis states since we deal with the 2-qubit circuit. The Fisher information matrix is calculated with $100$ uniform weights realization $\theta \in [0, 2\pi)$.

\begin{figure*}
    \centering
    \includegraphics[width=\textwidth]{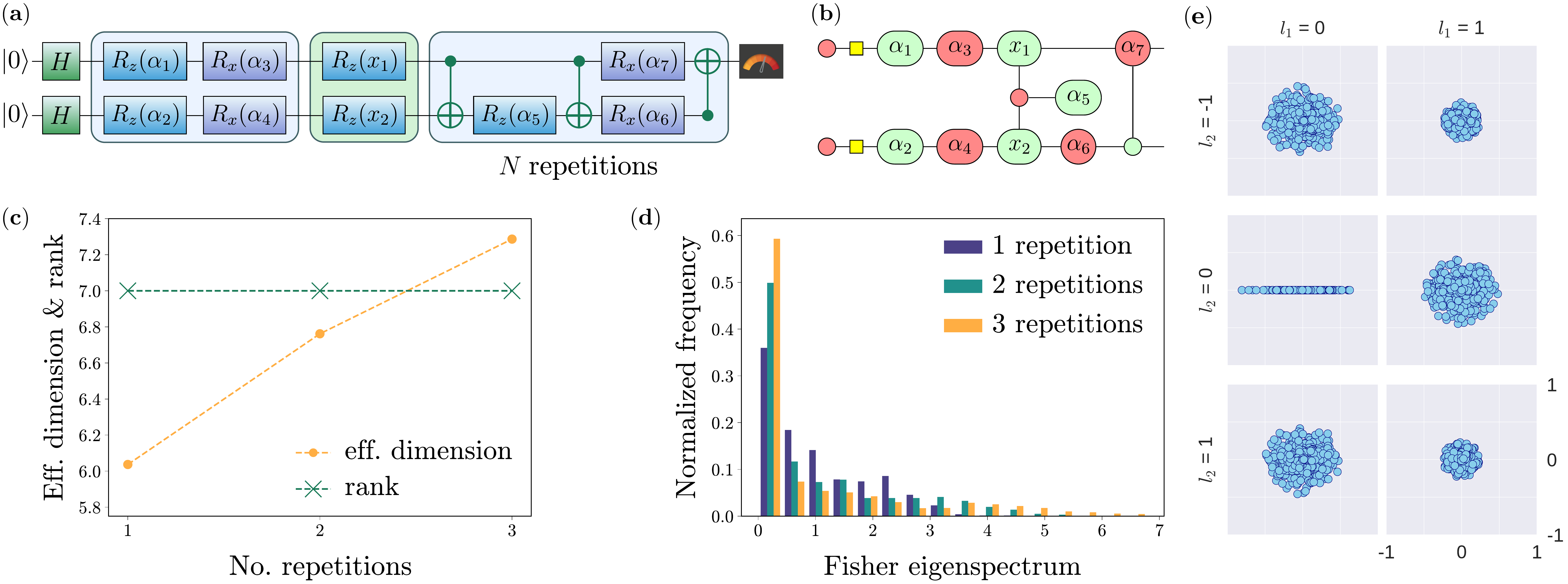}
    \caption{(a) Toy version of PQC used as a part of parallel HQNN in our study. (b) The ZX-calculus graph representation for PQC. No fundamental simplification; measurement is done in Z-basis, which ensures non-commutativity with previous gates. (c) Effective dimension and rank of the average Fisher information matrix for different repetitions $N={1, 2, 3}$ with $7, 10$ and $13$ trainable parameters correspondingly. Not maximal rank for networks with $N={2, 3}$ indicates the presence of zero gradient parameters. (d) Fisher information matrix normalized eigenspectrum frequency. The degeneracy about zero means lower trainability. (e) The Fourier accessibility. The set characterizes the possible values of the particular Fourier coefficient.
}
    \label{fig: analysis}
\end{figure*}

Figure \ref{fig: analysis}(c) shows how the average Fisher information matrix's effective dimension and rank depend on the network's trainable layers. As expected, the effective dimension increases with the number of trainable parameters, indicating an increase in expressivity. However, trainability is also an important factor. The spectrum of the Fisher information matrix reflects the square of the gradients \cite{Abbas2021Power}, and a network with high trainability will have fewer eigenvalues close to zero. 

Our experiments found that the Fisher information matrix rank remained constant at $7$ for all three configurations, indicating the presence of zero gradients for some network parameters with $10$ and $13$ trainable parameters. This is further illustrated in Figure \ref{fig: analysis}(d), which shows the distribution of eigenvalues for each configuration. The probability of observing eigenvalues close to zero increased from $36\%$ for one repetition to almost $60\%$ for three. Therefore, our results suggest that using only $N=1$ repetition is the optimal strategy for this setup.

\subsection{Fourier accessibility}\label{sec:appendix_fourier}

In Ref.~\cite{Schuld2021Effect}, it was demonstrated that any quantum neural network (QNN) can be expressed as a partial Fourier series in the data. The encoding gates in the QNN determine the frequencies that can be accessed. In the case of a multi-feature setting, the QNN produces a multi-dimensional truncated Fourier series. The quantum approximator $f(\boldsymbol{\theta},\mathbf{x})$, which is the expectation value of a specific measurement for a two-feature setting, can be expressed as a sum of truncated Fourier series terms:
\begin{equation}
     f(\boldsymbol{\theta},\mathbf{x}) = \sum_{l_1=-L_1}^{L_1}\sum_{l_2=-L_2}^{L_2} 2|c_{l_1,l_2}| \cos(l_1 x_1 + l_2 x_2 - \arg(c_{l_1,l_2})),
\end{equation}
where $L_1$ and $L_2$ are the numbers of encoding repetitions for the first and second features.
The Fourier coefficients, $c_{l_1,l_2}$, of the QNN determine the amplitude and phase of each Fourier term and depend on the variational gates used in the circuit. The amplitude of the coefficient is limited by the fact that the expectation value of any QNN takes values in the range of $-1$ to $1$. As a result, the maximum amplitude of $c_{l_1,l_2}$ is 1. The accessibility of the Fourier space for a QNN is evaluated by examining a family of quantum models with only two features and one encoding repetition. The circuit is set up in this case, and the weights are randomly varied many times.

The results of this analysis are shown in Figure \ref{fig: analysis}(e), which displays the Fourier accessibility of the network (with $N=1$ repetition) for $1000$ randomly generated weight sets in the range of $[0, 2\pi)$. The Fourier coefficients for a series with nine terms are presented, but due to the symmetry property $c_{l_1,l_2} = c_{-l_1,-l_2}$, only six coefficients are shown. It can be observed that the set characterizing the possible values of the coefficient does not degenerate into a point for any $l_1$ and $l_2$. Five out of nine coefficients have an amplitude of approximately or greater than $0.5$, while the other four have an amplitude of about $0.25$. Phase accessibility is also essential, and it can be seen that phases can be arbitrary except for $c_{0,0}$, which remains fixed. The Fourier accessibility shows comparable results with experiments conducted in Ref.~\cite{Schuld2021Effect}, which is sufficiently good.

\end{document}